\documentclass{webofc}
\usepackage[varg]{txfonts}   % Web of Conferences font
\usepackage{natbib}
\usepackage{algorithm}
\usepackage{algorithmic}
\usepackage{graphicx}
\usepackage{subfig}
\graphicspath{ {./images/} }
\begin{document}
\title{Accelerating GAN training using highly parallel hardware on public cloud}
%
% subtitle is optional

\author{\firstname{Renato} \lastname{Cardoso}\inst{1}\fnsep\thanks{\email{Renato.Cardoso@cern.ch}} \and
        \firstname{Dejan} \lastname{Golubovic}\inst{1} \and
        \firstname{Ignacio} \lastname{Peluaga Lozada}\inst{1}
        \and
        \firstname{Ricardo} \lastname{Rocha}\inst{1}
        \and
        \firstname{João} \lastname{Fernandes}\inst{1} 
        \and
        \firstname{Sofia} \lastname{Vallecorsa}\inst{1} 
        % etc.
}

\institute{CERN, 1 Esplanade des Particules, Geneva, Switzerland}

\abstract{%
  With the increasing number of Machine and Deep Learning applications in High Energy Physics,  easy access to dedicated infrastructure represents a requirement for fast and efficient R\&D. This work explores different types of cloud services to train a Generative Adversarial Network (GAN) in a parallel environment, using Tensorflow data parallel strategy. More specifically, we parallelize the  training process on multiple GPUs and Google Tensor Processing Units (TPU) and we  compare two algorithms: the TensorFlow built-in logic and a custom loop, optimised to have higher control of the elements assigned to each GPU worker or TPU core. The quality of the generated data is compared to Monte Carlo simulation. Linear speed-up of the training process is obtained, while retaining most of the performance in terms of physics results. Additionally, we benchmark the aforementioned approaches, at scale, over multiple GPU nodes, deploying the training process on different public cloud providers, seeking for overall efficiency and cost-effectiveness. The combination of data science, cloud deployment options and associated economics allows to burst out heterogeneously, exploring the full potential of cloud-based services. %These services, when validated and scaled up for the deployment of research use cases with optimised costs, establish the ground for a sustainable path to explore innovative workflows for data intensive science.
}
\maketitle
\section{Introduction}

 In recent years, several studies have demonstrated the benefits of using Deep Learning (DL) to solve different tasks related to data processing in High Energy Physics (HEP): for example, generative models are being tested as fast alternatives to Monte Carlo based simulation and anomaly detection algorithms are being explored to design searches for rare new-physics processes. 

Training of such models has been made tractable with the improvement of optimization methods and the advent of dedicated infrastructure.  High Performance Computing (HPC) and storage technologies are often required by these types of projects, together with the provision of high processing capacity on multi-architectures, ranging from large multi-core systems to hardware accelerators like GPUs or TPUs. %Memory consumption, for example, represents a limiting  factor  in  many  applications: millions of read-out channels of a detector produce an amount of data that is usually too large to be processed as a whole. The standard approach, in these cases, is to segment and select regions of interest for further processing. The possibility to analyse larger samples and, consequently, be able to process deeper models represents an advantage, both in terms of accuracy and computing resources exploitation.
These requirements directly translate into the necessity to extend the HEP computing model that, so far, has relied on  on-premise resources and grid computing. A hybrid model has the potential to seamlessly complement and extend local infrastructure, using widely supported techniques, allowing flexible and efficient resource bursting to public cloud for CERN research teams and science beyond.
In this context, our work explores the deployment of HEP DL models on native cloud-based frameworks, with well-established industry support and large communities, whilst, at the same time, integrate it with the ongoing work to transparently and generically burst CERN on-premises infrastructure to public clouds.
For this reason, in addition to the optimisation of the specific physics use case, we provide relevant feedback on the most efficient models to develop an end-to-end integrated strategy that includes cost prediction and optimization when consuming public cloud resources at scale, for scientific research.
The development and optimisation of deep generative models as fast simulation solutions is one of the most important research directions as far as DL for HEP is concerned.
HEP experiments rely heavily on simulations in order to model complex processes and describe detectors response: the classical Monte Carlo approach can reproduce theoretical expectations with a high level of precision, but it is extremely demanding in terms of computing resources \cite{hsf2019}. %The most important scientific results obtained at the LHC, so far, have relied on the large distributed infrastructure of the Worldwide LHC Computing Grid (WLCG) \cite{WLCG}: an infrastructure spanning 170 computing centers across the world and about a million computing cores. A large fraction of this computing power is dedicated to Monte Carlo simulations: as an example, the ATLAS experiment, one of the LHC experiments at CERN, quoted about 70 \% of its WLCG resources were dedicated to simulations in 2018 \cite{Atlas_chep2018}. 
For this reason, different strategies trading some physics accuracy for speed are being developed as fast simulation techniques and deep generative models are today among the most promising solutions \cite{GAN1,GAN2,WGAN_Thorben,Salamani2018}.\\
This work focuses on 3DGAN \cite{GAN_Gulrukh}: a three-dimensional convolutional Generative Adversarial Network (GAN) that is being optimised for the simulation of electromagnetic calorimeters.
This work summarises the specific development required by the 3DGAN architecture and the adversarial training process for efficient data parallel training on Google TPUs and multiple GPUs setups. It also details the effect of distributed and mixed-precision training (on TPUs) on  3DGAN physics performance: this aspect is essential to ensure that the performance of complex simulations tasks (such as the GAN generation process) is preserved at an acceptable level. %In most cases so far, strategies aimed at accelerating the training process, such as distributed training or reduced precision data representation, have been benchmarked on binary or multi-label classification tasks \cite{classification_quantization}, which present requirements (i.e. a classification accuracy unchanged within a few percent) that are relatively simple when compared to the typical requirements of generative models in scientific applications, where a high level of fidelity is expected for the synthetic data.
In addition, we describe the 3DGAN training deployment on public cloud providers (Google Cloud Platform (GCP) and Microsoft Azure) comparing different approaches, based on Kubernetes, Kubeflow and the Azure Machine Learning service. Our deployments are among the first examples of establishing an innovative hybrid platform successfully deploying scientific deep learning workloads on public clouds at large scale \cite{kube}.
%Our paper is organised as follows: section \ref{sec:pr} contains a brief review of related work in the HEP domain. Section  \ref{sec:3dgan} and \ref{sec:training} briefly describe the 3DGAN model and the optimisation of the adversarial training process. The results obtained on TPUs in terms of training time scaling and physics performance follow. The second part of the paper describes the deployment strategies adopted both on GCP and Microsoft Azure, discussing the 3DGAN behavior on different multi-GPU hardware configurations considering overall efficiency and cost-effectiveness when using cloud-based resources. We conclude the paper summarizing our findings and giving an outlook into future work.

\section{Generative Adversarial Networks in High Energy Physics}
\label{sec:pr}
Generative models represent a fundamental part of deep learning and
%Through the years this field has seen developments in the Generative Stochastic Networks~\cite{DBLP:journals/corr/BengioT13}, to the Variational AutoEncoders~\cite{2013arXiv1312.6114K}, and Generative Adversarial Networks (GAN)~\cite{2014Goodfellow}. 
GANs~\cite{2014Googfellow}, in particular, can generate sharp and realistic images with high resolution \cite{wgan,karras2017progressive}. %Inspired by the game theory, adversarial training is defined as a competition between two players: a generator and a discriminator. The discriminator distinguishes real from fake images while the generator tries to fool the discriminator by producing an output as realistic as possible. The process eventually results in the generator learning the distribution of the real data if given enough representation capacity and time. 
%The 3DGAN architecture is inspired by Auxiliary Classifier GANs (ACGAN)~\cite{acgan}, a natural extension of the GAN approach that features a faster, more stable convergence by introducing auxiliary tasks for the discriminator. ACGAN is a simple architecture and, at the same time, the most suitable configuration for the task at hand: application to simulation requires, indeed, generating images conditioned on a set of inputs. Therefore, introducing auxiliary tasks, such as regression on the conditioning variables,  not only stabilizes the training but also provides a feedback on the quality of the conditioning. 
In HEP, most GAN applications address the problem of calorimeters simulation.
The LaGAN~\cite{de2017learning} and CaloGAN~\cite{paganini2017calogan} models first introduced GAN for simulating simplified detectors; since then, multiple prototypes have been developed~\cite{Salamani2018,WGAN_Thorben,icip18_3dgan}. %The 3DGAN architecture is inspired by Auxiliary Classifier GANs (ACGAN)~\cite{acgan}, a natural extension of the GAN approach that features a faster, more stable convergence by introducing auxiliary tasks for the discriminator.
%While most of those studies focus on specific detectors, the 3DGAN prototype investigates techniques for generalising to multiple detector geometries by optimising the GAN architecture through hyper-parameter searches. As such, it entails reducing training time through distributed techniques and optimisation for dedicated hardware accelerators.
While distributed deep neural networks training is a well-established field, with multiple approaches thoroughly investigated and state-of-the-art solutions widely spread, the specific GAN use case is still a relatively new research subject:  \cite{mustafa2019cosmogan} represents a notable example at large scale, while additional work can be found in \cite{IXPUG_3DGAN,blind3_3dgan}.
Low precision computation (int8, float16, bfloat16) is another active research field, aimed at reducing computing resources, mostly targeted to speed-up inference: models are trained in float32 precision, on multi-node clusters, quantized to lower precision and then loaded on light-weight devices such as edge-devices.
%actively researched in the recent years to allow, in particular, fast inference on mobile devices where the computational power is strongly constrained, but more generally, with increasing model size and complexity, constraints on computing time and memory consumption become relevant for other architectures, CPUs and accelerators alike.
Lower precision formats can be used at training time as well, in combination with full precision float32 in order to maintain the level of accuracy (mixed precision training) \cite{MixedPrecision2,IntelMixedPrecision}.

%Need related work on  cloud deployment

%\section{The 3DGAN model}
%\label{sec:3dgan}
%HEP calorimeters output can be interpreted as a three-dimensional grid in which each pixel represents a single sensor unit: the overall detector response can therefore be regarded as a 3D image in which each pixel represents the intensity of a specific physics measurement.  %Typical computer visions techniques, such as convolutional layers, can then be used to analyse or simulate the detector data. 
%However, detector "images" exhibit specific features that strongly differentiate them from typical RGB images: they are extremely sparse (most of the pixels are empty) and pixel intensities have wide dynamic range spanning several orders of magnitudes. 
%As an example, figure \ref{GANevent} shows the three projections, along the canonical axes, of a calorimeter 3D image.
%\begin{figure}
%    \centering
%    \subfloat\centering %{{\includegraphics[width=\textwidth]{images/GANevents.png} }}
%    \caption{Two dimensional projections on the $xy$, $xz$ and $yz$ planes %f an example calorimeter three-dimensional out, generated by 3DGAN\cite{GAN_Gulrukh}. The color scale represents the pixel intensity (i.e. the particle energy measured by the detector sensor at that position on the grid).}
%    \label{GANevent}
%\end{figure}
Details about the 3DGAN model can be found in \cite{GAN_Gulrukh}: it represents the first proof of concept on using 3D convolutional GANs to simulate high granularity calorimeters. 
The 3DGAN model uses an ACGAN-like~\cite{acgan} approach to introduce physics constraints and improve image fidelity: in particular, two quantities, representing the particle energy, $E_{P}$, and its trajectory, measured by the angle $\theta$, are used as input to the generation step. %The novelty of our works resides in the high granularity (high spatial resolution) of the detector we simulate and in the use of three dimensional convolutions that are essential to preserve all spatial correlations between pixels. Lastly our particular loss function and preprocessing results in high accuracy for a more complex scenario as compared to other GAN applications for HEP calorimeter simulation, involving a wider range of input variables used for conditioning.
%The 3DGAN consists of a discriminator ($D$) and a generator ($G$) network. 
  %The 3DGAN generator, with seven 3D convolutional layers, is stronger than the discriminator, with four layers: batch normalization is applied to stabilise training, while a combination of Relu~\cite{relu} and leakyRelu~\cite{leakyrelu} activation  functions are used throughout the architecture in order to induce the high level of sparsity expected in the detector data. The discriminator is regularized by using a $20\%$ dropout and a single average pooling layer (additional pooling layers result in substantial performance loss).
%The discriminator network has two trainable outputs: $O_{GAN}$ estimates the GAN real/fake probability via a sigmoid neuron and $O_{E_P}$ predicts the input particle energy as an auxiliary regression task (similar to~\cite{acgan}) through a linear neuron. Two additional outputs, $O_{E_{CAL}}$ the total  energy measured in the detector,  and $O_{\theta}$, an angle describing the input particle trajectory, are implemented as lambda layers and  represent non-trainable physics-based constraints. 
In general terms, simulating samples using generative models is much faster than using a Monte Carlo approach. In the case of 3DGAN, a several orders of magnitude speed-up is observed with respect to the HEP Monte Carlo state-of-the-art \cite{geant4}: a 67000x speed-up has been recently achieved using int8 quantized inference on Intel Xeon Cascade Lake \cite{icipram}. 
The training process is however very time consuming: an entire week is needed in order to train the model  to convergence using a single NVIDIA V100 GPU and therefore a distributed training approach is essential \cite{CHEP2018_distributed}. %In order to reduce the training time we have interfaced 3DGAN to several distributed frameworks, including Horovod \cite{horovod} and mpi-learn \cite{mpi-learn}, and benchmarked the parallel training process on different HPC systems \cite{IXPUG_3DGAN, CHEP2018_distributed}. 
\section{The adversarial training process acceleration}
\label{sec:training}
%\subsection{Tensorflow distributed training strategies}
The Tensorflow 2.3 {\it distribute.Strategy} API  runs the training process across a single node with multiple GPUs, multiple nodes, as well as multiple TPU cores, while ensuring portability with minimal code changes; the downside is that only graph mode is available on TPUs, whereas time penalties are introduced in eager mode on GPUs. This work relies on the {\it mirrored strategy}, that implements synchronous parallel training over multiple GPUs on a single node; the {\it multi-worker mirrored strategy}, which extends it across multiple nodes and the {\it TPU strategy}, that implements synchronous training on TPUs, using its own, optimised, {\it All\_reduce} and collective operations \cite{tensorflow2015-whitepaper}. 
%\subsection{Strategies}
%Tensorflow offers a number of different strategies depending on the type of architecture we are distributing our model. In our case we decided to use 3 different strategies:
%\begin{itemize}
%\item{ Any variable or model created within its scope is replicated across all GPUs, the forward propagation and the loss and gradients calculations  are done on each GPU and only then an {\it All\_reduce} algorithm is applied to sync up the gradients and update all GPU models equally, thus limiting the number of communication steps}

%\item{ , by introducing a Collective Operation that is able to automatically choose the best {\it All\_reduce} algorithm and apply communication optimizations depending on the system topology. This strategy uses an environment variable, TF CONFIG, to configure the cluster, set up the worker machines, and distribute tasks.}

%\item{ The TPU cluster resolver needs to be previously initialized with the address of the TPU in the google cloud platform.}
%\end{itemize}

The 3DGAN adversarial training, shown in Algorithm \ref{alg0}, trains, alternately, the discriminator and generator networks an equal number of times. It includes several steps to initialise the generation process, train the discriminator network on real and fake data and, finally, the generator itself. %The RMSprop optimiser is used.

\begin{algorithm}
\scriptsize
\caption{\scriptsize The 3DGAN training process}
\label{alg0}
\begin{algorithmic} 
\FOR{$N_{epochs}$}
\FOR{$N_{batches}$}
\STATE get 3D image batch
\STATE get labels: $E_p$, $E_{CAL}$, $\theta$
\STATE generator input: sample latent noise batch and concatenate(noise, $E_p$, $\theta$)
\STATE generate fake 3D image batch
\STATE calculate fake $E_{CAL}$ batch
\STATE train discriminator on real batch
\STATE train discriminator on fake batch
\FOR{2}
\STATE get labels: $E_p$, $E_{CAL}$, $\theta$
\STATE generator input: sample latent noise batch and concatenate(noise, $E_p$, $\theta$)
\STATE train the generator
\ENDFOR
\ENDFOR
\ENDFOR
\end{algorithmic}
\end{algorithm}
 
\begin{figure}
    \centering
    \subfloat\centering {{\includegraphics[width=0.4\textwidth]{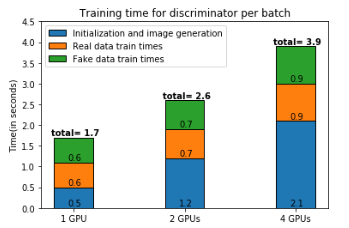} }}
    %\qquad
    \subfloat\centering {{\includegraphics[width=0.4\textwidth]{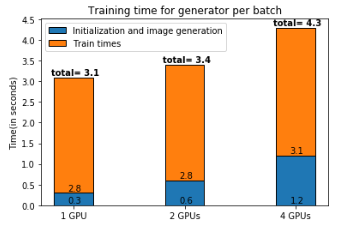} }}
    \caption{(left) Discriminator training times divided between processing, real data training and fake data training. (right) Generator training times. Times are quoted for one data batch.}
    \label{Version1}
\end{figure}

Algorithm \ref{alg0} can be implemented using  standard training APIs, such as {\it keras.train\_on\_batch} or custom training loops. 
In the first case, the distribution logic is built-in and only the batch size is manually updated according to the number of parallel replicas. Taking a look at the pseudo-code in Algorithm \ref{alg0}, it is evident that the generator input initialization steps are not distributed by {\it keras.train\_on\_batch}: they are sequentially run and end up being a bottleneck when the number of replicas increases. This effect is shown in Figure~\ref{Version1}: the discriminator training time (per batch) stays the same while increasing the number of GPUs, with a small increase due to reduce and broadcast operations. On the other hand, the generator initialisation time increases linearly with the number of GPUs, since it is sequentially run. 

%This approach, though easy to deploy, has a major disadvantage when applied to GAN training, as it is not suitable to parallelise all training steps, which then result in a bottleneck. 
%\subsubsection{Second approach custom training loop:}
%\hfill \break
%\hfill \break
In order to reduce this bottleneck, we rewrite Algorithm \ref{alg0} as a custom training loop, using Tensorflow {\it tf.function} to explicitly redefine the forward step in training mode. % model in training mode, the loss calculation, using {\it keras.loss} primitives, the gradients calculation and finally the weights update using the RMSProp otpimizer, which also synchronises the gradients among all replicas.
%that is used to run the training steps, the problem with using a training function is that, it limits what can be done inside itself, for once you can't use eager mode inside a tf.function and more importantly you can't use a tf.function inside another tf.function.
%The inability to use a tf.function inside one, makes us unable to use the train on batch function, because this is a tf.function, meaning we need to redefine the behaviour of the train on batch function. The train on batch has 4 steps, a forward pass, the calculation of the loss, the calculation of the gradients and the updates of the weights using an optimizer. This is done by calling t
In addition, the {\it tf.function} includes all previously sequential steps.  
%Need details on 
%\begin{figure}%
%    \centering
%    \subfloat\centering %{{\includegraphics[width=0.45\textwidth]{Version2_train_times.png} }}
    %\qquad
%    \subfloat\centering %{{\includegraphics[width=0.45\textwidth]{Version2_times_vs_linear.png} }}
%    \caption{(left) Train times per batch for the training step. (right) Comparison between the measured training time and a linear speed-up, %per epoch.}
%    \label{Version2}
%\end{figure}
%Figure~\ref{Version2} shows the corresponding improvement, with batch %training time that stays constant while increasing the number of GPUs. Also %shown is the total time per epoch which decreases almost linearly up to 8 %GPUs on a single node: 
At this point, the remaining bottleneck is represented by the  data pre-processing and distribution steps. In order to address this issue, we rely on some important advantages related to using a customised training loop: loading data directly to the accelerators (GPUs and TPUs) and running data preparation (batching and shuffling) on the CPU host while the GPUs/TPUs are training. 
More specifically, we convert the original HDF5 data to the native tensor format, TF Records, and introduce iterators instead of manually instantiating batches on the CPU host. This final optimisation step yields results we discuss in the next section for TPUs and in section \ref{sec:cloud} for a larger number of GPUs on Google Cloud Platform (GCP) and Microsoft Azure clouds.

\section {Training on Google Tensor Processing Units}
\label{sec:tpu}
%Due to the increasing need for faster machine learning approaches, ML/DL specialized hardware configurations and Domain Specific Architectures (DSA) are becoming essential.
%Common approaches use distributed environments with ML targeted  GPUs, such as  NVIDIA V100, with specialized topologies and dedicated interconnects for faster data distribution.
Application-specific integrated circuits (ASICs), such as Google Tensor Processing Units (TPUs) dedicated to ML/DL workloads, are being developed with the objective of accelerating matrix multiplication.
%When working with multiple chip architectures~\cite{10.1145/3360307} it is important to take into account how these chips communicate with one another. this is for example much easier to do for DSAs, for which communication patterns are known.%: more specifically, in this case, we know we need to reduce all the weights and update them across all the nodes. 
%\subsection {TPUs Hardware Configuration}
In order to have efficient communication across TPUs, Google implements a 2D Torus topology by using Inter-Core Interconnect (ICI) links that enable a direct connection between chips and simplify the rack-level deployment. In addition, the v2 and v3 TPUs use two cores per chip, further decreasing the latency with respect to single-core chips. In particular, each TPU core contains a 128x128 matrix multiplier unit (MXU) which uses systolic arrays. 
%Each TPU core  contains:
%\begin{itemize}
%  \item  A 128x128 matrix multiplier unit (MXU) which uses systolic arrays. 
%  \item An ICI responsible for the communication between chips.
%  \item An High Bandwidth Memory (HBM) to reduce memory bottleneck.
%  \item The Core Sequencer that fetches very long instruction words and software-managed memory instructions, executes scalar operations and forwards vector instructions to the Vector Processing Unit.
%  \item The Vector Processing Unit (VPU)
%  \item A Transpose Reduction Permute Unit responsible to do matrix transposes, reduction and permutations of the VPU lanes.
%\end{itemize}

%\subsection{Bfloat16 and XLA}
Google TPUs use bfloat16 for MXU calculations, in order to accelerate data processing while, at the same time, maintaining a high level of accuracy. %The bfloat16 format keeps the same exponent of a 32-bit float while decreasing the mantissa, this way preserving small updates values without sacrificing speed.
To fully exploit bfloat16, the Tensorflow XLA compiler maps TPU dependent instructions while also targeting CPU and GPUs instructions.
The XLA compiler analyses the code and optimises the data transfer among different TPU elements, maps concurrent instructions in different parts of the TPU hardware and increases the ICI communication capabilities between TPU cores. All these optimisation steps occur during the first {\it tf.function} iteration, which in consequence, is slower. %This effect is evident in the first batch iteration that takes around 5 minutes to compile and generate the XLA code, while every subsequent training step takes considerable less time. 
For this reason we quote average results excluding the first batch.

\subsection{Results}
\begin{figure}%
    \centering
    \subfloat\centering {{\includegraphics[width=0.3\textwidth]{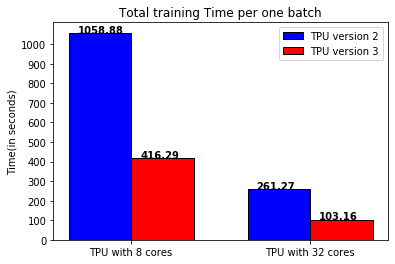} }}
    %\qquad
    \subfloat\centering {{\includegraphics[width=0.3\textwidth]{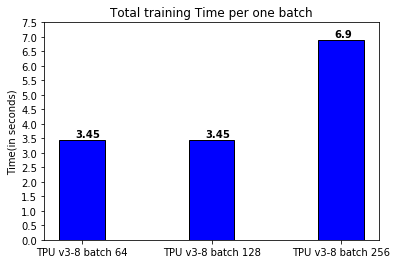} }}
     \subfloat\centering {{\includegraphics[width=0.3\textwidth, trim={0.25cm 0.6cm 0.5cm 0.4cm}, clip=true]{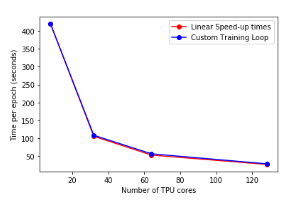} }}
    \caption{(left) Time per epoch (in seconds) on  8 cores v2 and v3 TPU, $BS=128$. (center) Time per epoch (in seconds) on 8 cores v3 TPU for batch sizes $BS=64,128,256$. (right) Measured training time per epoch (in seconds) compared to the theoretical linear speed-up.}
    \label{TPUv3}
\end{figure}

\begin{figure}[H]
    \centering
   % \subfloat\centering {{\includegraphics[width=5.6cm]{TPU_times.png} }}
   % \qquad
     \subfloat\centering {{\includegraphics[width=0.3\textwidth,trim={1.8cm 8.6cm 2.25cm 9.35cm}, clip=true]{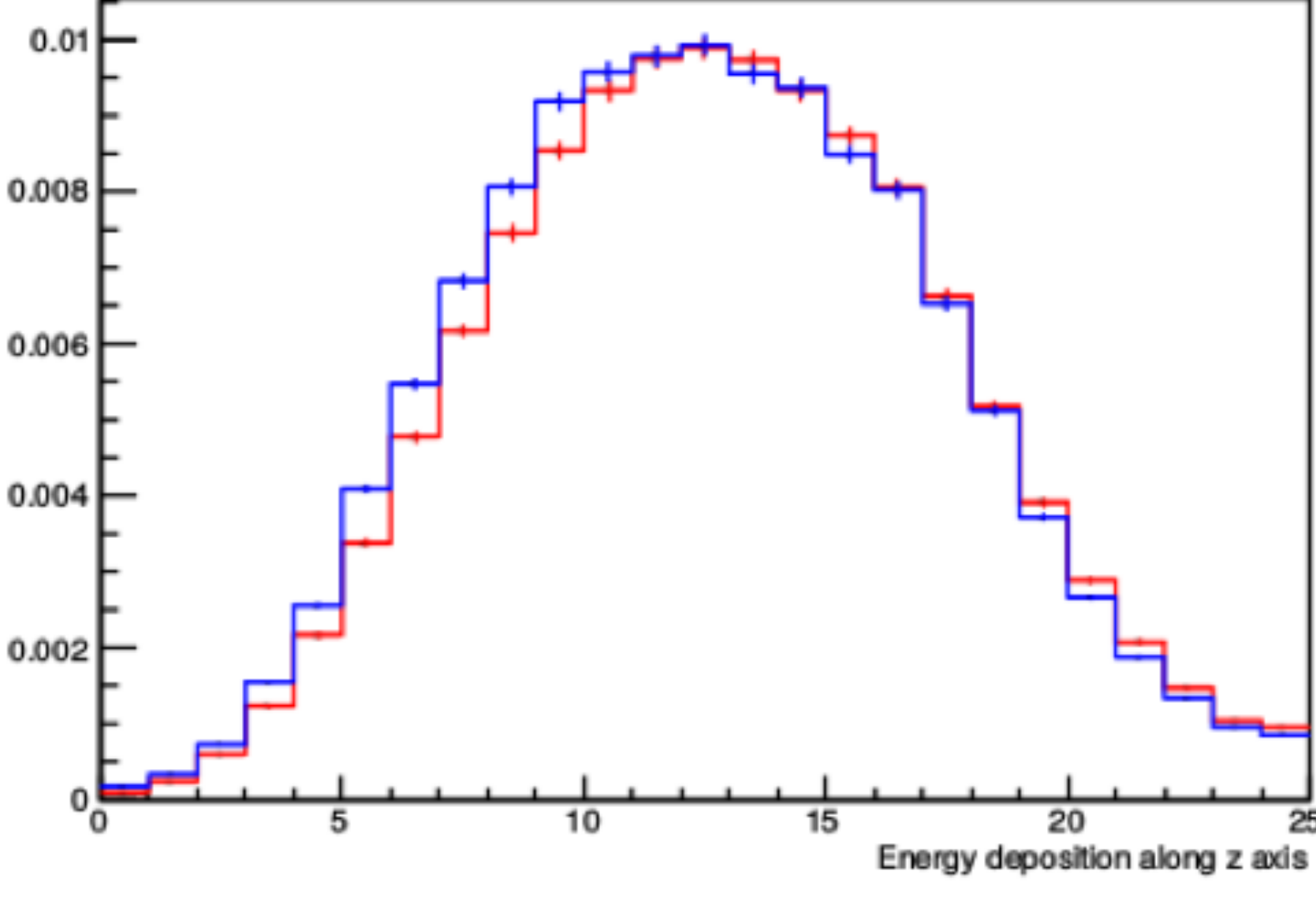} }}
     \subfloat\centering {{\includegraphics[width=0.3\textwidth, trim={2cm 9cm 2cm 9cm}, clip=true]{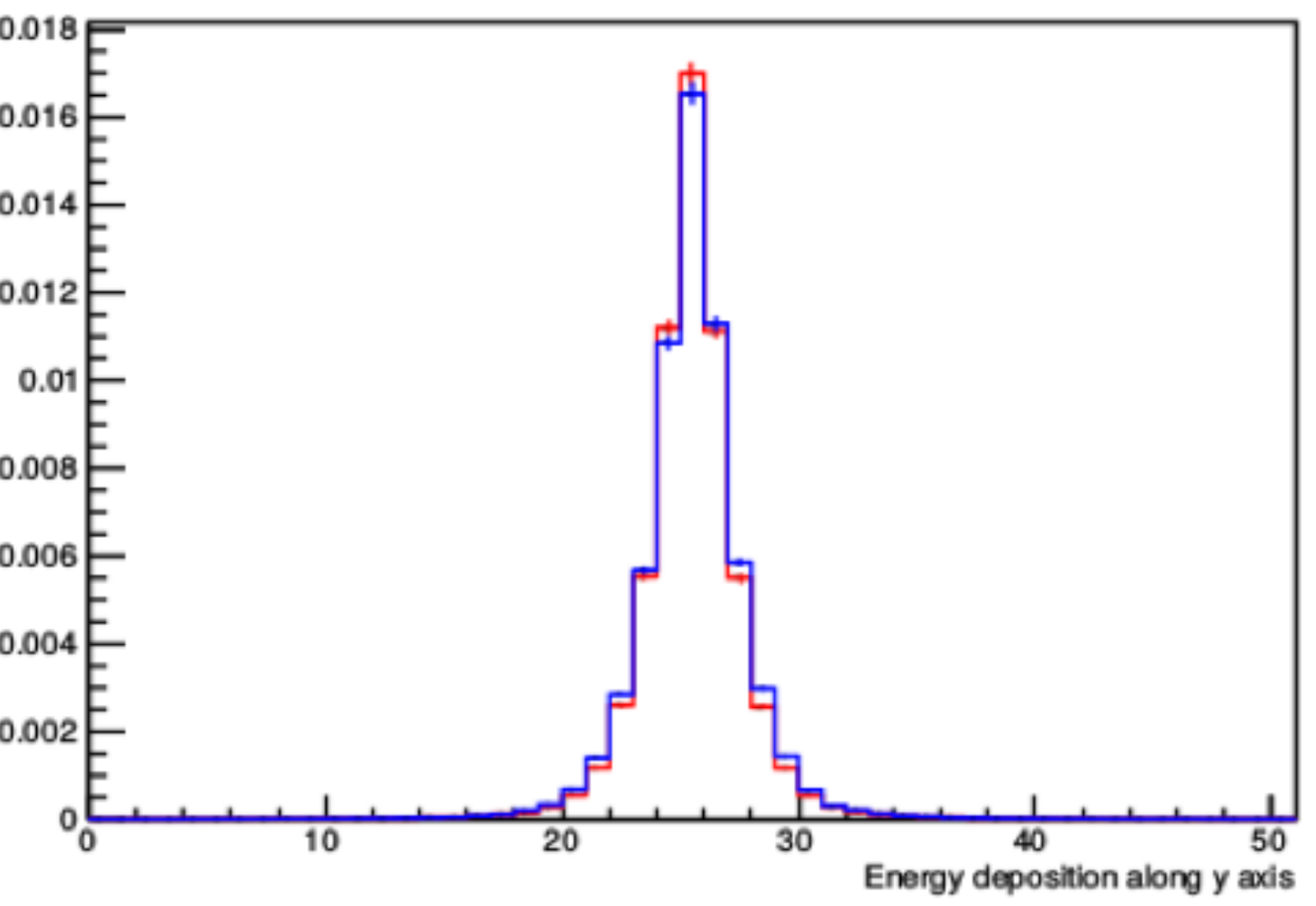} }}
    %\qquad
     \subfloat\centering {{\includegraphics[width=0.3\textwidth,trim={4cm 3cm 4cm 3cm}, clip=true]{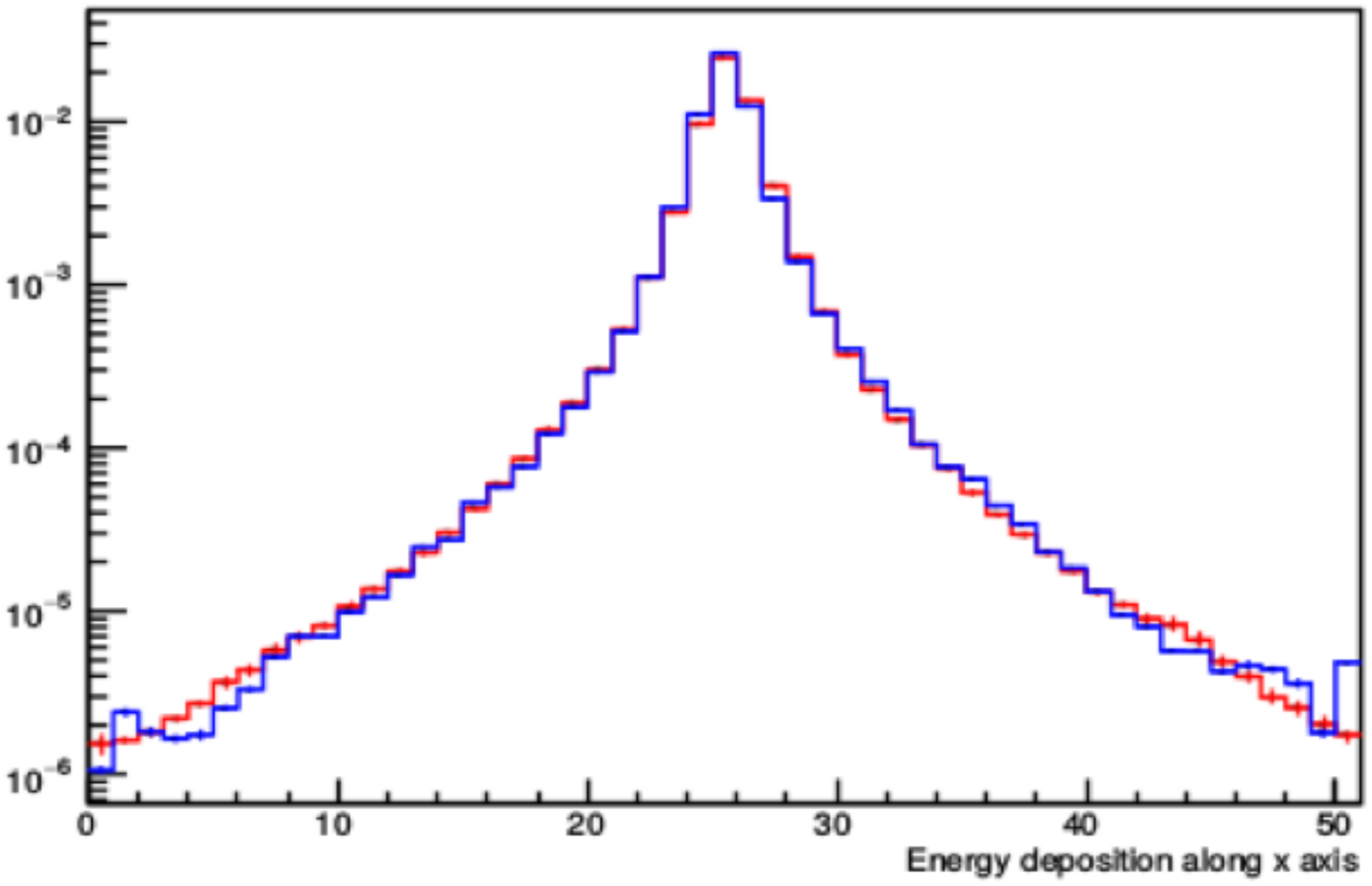} }}
    \caption{Calorimeter energy response. The GAN prediction is in blue, Monte Carlo in red. (left) Along the detector longitudinal plane. (center) Along the detector transverse plane in linear scale and (right) logarithmic scale.}
    \label{TPUv3_times}
\end{figure}

Our initial test is run on 8 cores TPUs, using a batch size $BS=128$ in order to match the design MXU dimension. Results for v2 and v3 TPUs are compared in  figure~\ref{TPUv3}: using the double core v3 TPU we obtain 2x speed-up on the epoch training time. Figure~\ref{TPUv3} also shows, on the center panel,  the batch size impact on training time on 8 cores v3 TPU: a batch size $BS=64$ is under utilising the MXU capacity and, therefore, it is processed in the same amount of time as a 128 batch; on the other hand, $BS=256$, which is double the optimal size, is processed using twice the time.  
Figure~\ref{TPUv3}, right panel, shows weak scaling results, obtained  using a $BS=128$ batch size on v3 TPUs with a number of cores ranging from 8 to 128: the time per epoch scales linearly. Running on a 128 cores TPU, reduces the training time per epoch to about 30 seconds, allowing 3DGAN to train to convergence in a little more than one hour.
As mentioned above, preserving the fidelity of the synthetic images is critical, when accelerating training. In order to verify this constraint, we validate the quality of the generated images against Monte Carlo simulation: the left panel on figure~\ref{TPUv3_times} shows the calorimeter energy response along the longitudinal plane. It can be noticed, that while the GAN model seems to slightly distort the response toward the left side of the distribution, the agreement remains overall very good.
The energy response along the transverse plane, shown in the center panel of figure ~\ref{TPUv3_times}, is also optimally reproduced, both in the central calorimeter cells (the distribution bulk highlighted in the linear scale plot) and at the volume edges (shown in the logarithmic scale plot). %No performance degradation is observed due to the use of bfloat16 at training time.
%As an initial benchmark we run our approach with different TPU setups, without changing the training parameters,  to see how these influence the times obtained, after that we used the optimal setup and increased the number of TPUs obtaining the times and speedup for the distributed training.

To further analyse the results obtained, a profiling tool was employed using tensorboard. %As we can see in figure~\ref{TPU_profile} 
The program is mostly compute bound, in fact the TPUs are only idle 0.7\% of the time, with 38\% of the time being used for the forward-pass (represented as the functional models), 61\% for back-propagation (the gradient tapes) and  0.2\% for {\it All\-Reduce}. 

%\begin{figure}[H]
%    \centering
%    %\subfloat\centering {{
%    \includegraphics[width=\textwidth, trim={0cm 1cm 0cm 0cm}, clip=true]{images/TPU_profile.png} %}}
%    \caption{Tracing profile of a single TPU core for a batch of size 128, run using 128 v3 TPU cores}
%    \label{TPU_profile}
%\end{figure}

On the other hand, analysis on the GPUs shows an increase on the idle time to around 2.9\%, for a higher number of GPUs, this is due to an increase in the time it takes to do the {\it All\-Reduce} operation, which occurs after every gradient tape operation shown in the profile.   

%Firstly we run our training in TPUs v2 and TPUs v3, we used a batch size of 128 on both of them and obtained the average training times for one epoch, see figure~\ref{TPUv3}. One of the big disadvantages of TPU v2 is that it only has a single MXU per core, which substantially increases the time it takes for a training step, more or less 2.5 times more from our results. With this we decided to use TPU v3 for any subsequent training measurement and tests we did.

%Secondly we tested what was the impact of different batch sizes for the TPU version 3. by keeping the same number of cores, 8 cores per TPU, see figure~\ref{TPUv3}. The time to do a batch of 64 and a batch of 128 is the same, this is due to the size of the MXU, being it a matrix of size 128 by 128, so to use it optimally a minimum batch size of 128 is necessary, any less the TPU will take the same amount of time without having any improvement. The same logic applies to a batch size of 256, as we are using twice the optimal batch size, it will end up taking twice the time to process it, in this case we don't see any improvement for using a batch size of 256. With this we decided to use a batch size of 128 as this shows the best results.
 %We then proceeded to run our test on multiple TPUs, see , we run on all the available TPU configurations beginning with 8 and ending with 128. We verify that our approach is able to maintain a linear speedup when increasing the number of TPUs.

\section{Automated deployment  using public cloud services}
\label{sec:cloud}

Training deep generative models for simulation is a workload that fits well the public cloud for multiple reasons. First of all, training is done periodically and can greatly benefit from the use of on-demand resources when a longer-term investment on-premises cannot be justified. Secondly the availability of a large number of GPUs, often a scarce resource on data centers on-premises, as well as access to the most recent architectures. Finally, the access to specialized accelerators such as Google TPUs and Graphcore IPUs, that are vendor specific, allowing for a generic unlocked deployment strategy, whilst at the same time, benefiting from a side-by-side comparison.

The deployment can take different approaches, including the low level management of virtual machines which keeps the burden of a significant fraction of the infrastructure management on the end user; on the other hand, one can also use open platforms such as Kubernetes, which offer APIs abstracting the infrastructure and delegating most of the operations to the cloud service itself, being available in all major public clouds; another possibility is to use MLaaS provided by the vendor, that normally consists in a PaaS optimised and dedicated to machine learning, such as the Azure Machine Learning service.
Below, we describe and present results for the last two approaches.

\subsection{Google Cloud Platform with Kubeflow}

%\subsection{Infrastructure Setup}

Our first setup relies on Kubeflow \cite{kubeflow}. Our existing on-premises deployment and configuration was reused and directed to the Google Kubernetes Engine (GKE), Google Cloud's managed Kubernetes service offering, benefiting from the resources under CERN IT's Cloud Broker Pilot project (CloudBank EU)\cite{CloudBank}. To monitor resource usage during training we rely on Prometheus and custom Grafana dashboards showing GPU power usage, memory usage, temperature and overall utilization.

The Kubernetes cluster includes multiple node groups, each of them with a profile where the number of GPUs per node varies from 1 to 8. This allows the evaluation of the penalty introduced from spreading the training across a larger number of nodes, as well as the penalty of increasing the density in a single node. 
Instances are of the n1-standard family, from n1-standard-4 for single GPU nodes to n1-standard-32 for 8 GPU nodes.
All tests rely on preemptible instances, which are low SLA resources living for a maximum of 24 hours that can be deleted at any point if capacity is reclaimed - but are over 3 times cheaper compared to reserved instances for V100 GPUs. No reclaim was experienced during the full set of tests performed.

%\subsection{TensorFlow Jobs}

%Kubeflow offers multiple custom resources to wrap distributed and non-distributed training in a single definition, supporting popular machine learning frameworks. A TFJob is used for TensorFlow and offers a way to integrate with the full set of TensorFlow capabilities. In particular it simplifies distributed training by providing a scheduling mechanism and automatically setting the configuration of the workers. %Information about the workers in the cluster and the current worker is wrapped in an environmental variable TF\_CONFIG.
%The prerequisite for running a TFJob is a Docker image containing all the dependencies and the code. The job definition is done in a yaml file where users can select the number of replicas for workers, chief workers or parameter servers, removing the need to hard-coded them in the code. For every worker, additional information can be supplied, such as the image from which the script runs, command to run in the container, resources that the worker consumes and additional volume mounts.

\subsubsection{Experiments and Results}
%\hfill \break
%\hfill \break
%\subsubsection{Batch size}
%\hfill \break
%\hfill \break
In the initial set of runs, the goal is to determine an optimal batch size for training the neural network. %Batch sizes represent the number of examples which are processed by GPUs in one training step (one forward and backward pass). 
Higher batch sizes mean less steps to complete the entire data set, which results in faster completion of the training. Though smaller batch sizes can yield to better generalization performance, in this case we tolerate a small loss of generalization performance (as discussed in section \ref{sec:azure}). From the hardware perspective, higher batch sizes offer better utilization of GPUs by exploiting advantages of the hardware. For the purpose of the paper, as models are being trained on enterprise machine learning platforms, it is important to provide maximum utilization of resources in order to minimize the cost for training jobs. 
This experiment is conducted by fixing the number of nodes (4), GPUs per node (8) and the number of epochs (5). The only hyper-parameter varying is the batch size. Batch sizes are selected to be multiples of 2 due to the GPU Single Instruction Multiple Data (SIMD) organization. Minimum batch size is 16, since it is sufficiently small to provide good generalization performance. The maximum batch size (96) is selected to fit the GPU memory.
%\begin{figure}%
%\setlength{\belowcaptionskip}{-15pt}
%    \centering
%    \subfloat\centering {{\includegraphics[width=5cm]{images/batch_size.png} }}
%    %\qquad
%    \subfloat\centering {{\includegraphics[width=5cm]{images/n_workers.png} }}
%    \caption{(left) Impact of batch size on time to perform one epoch of training. (right) Impact of %various combinations of hardware layout and TF worker configuration on time to perform one epoch of %training. Orange represents the first set of runs, where hardware configuration is fixed with 4 nodes %with 8 GPUs each, while number of workers and GPUs per worker change. Blue represents the second set %of runs where number of workers and GPUs per worker correspond to number of nodes and GPUs per node.}
%    \label{kf-results-batch-size}
%\end{figure}
As expected, the training times are reduced when using larger batches. The increase in performance is prominent between batch size of 16 and 32. With batch sizes of 64 and 80 the difference is almost negligible.

\begin{figure}%
    \centering
    \subfloat\centering {{\includegraphics[width=0.3\textwidth]{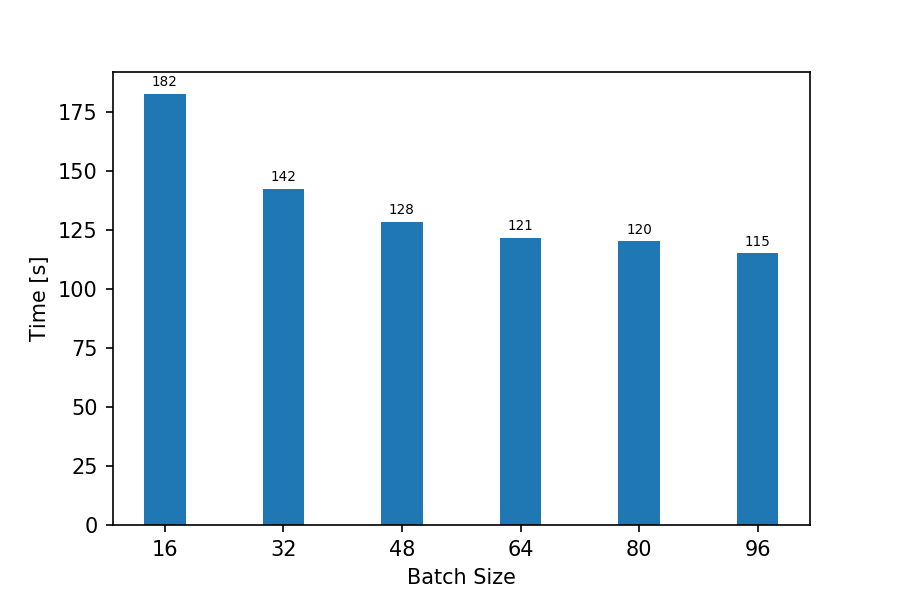} }}
    %\qquad
    \subfloat\centering {{\includegraphics[width=0.3\textwidth]{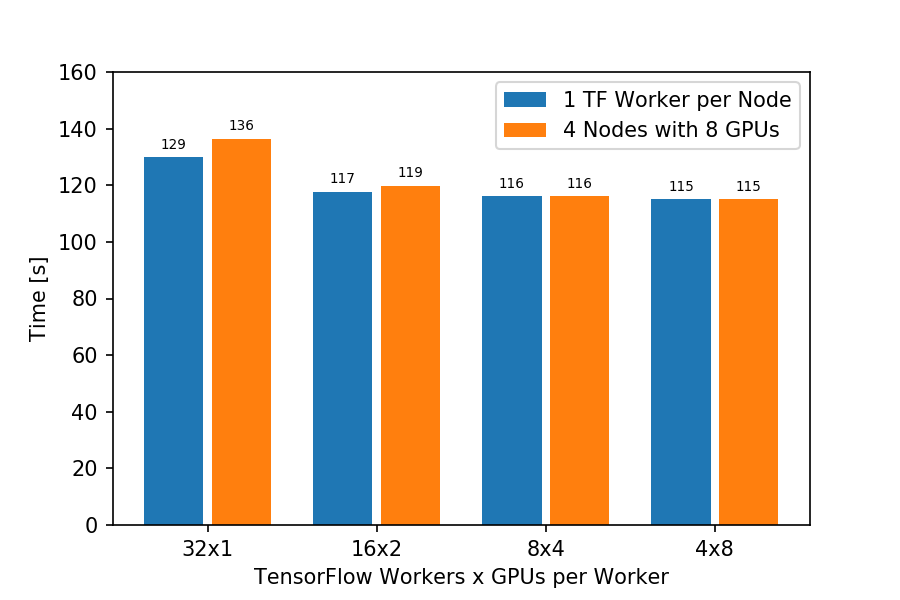} }}
     \subfloat\centering {{\includegraphics[width=0.27\textwidth, trim={0.25cm 0.4cm 0.5cm 0.4cm}, clip=true]{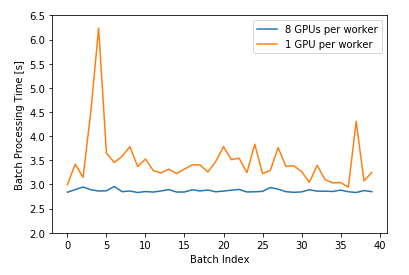} }}
    \caption{(left) Impact of batch size on time to perform one epoch of training. (center) Impact of various combinations of hardware layout and TF worker configuration on time to perform one epoch of training. Orange represents the first set of runs, where hardware configuration is fixed with 4 nodes with 8 GPUs each, while number of workers and GPUs per worker change. Blue represents the second set of runs where number of workers and GPUs per worker correspond to number of nodes and GPUs per node. (right) Batch processing times for the second epoch of training using hardware layout with 4 nodes with 8 GPUs per node. Orange represents the batch processing times when number of workers is 32 and each worker has 1 GPU, Blue represents the batch processing times when number of workers is 4 and each worker has 8 GPUs.}
    \label{kf-results-batch-size}
\end{figure}

%\hfill \break
%\hfill \break
%\subsubsection{Cloud Infrastructure Layout and TensorFlow Worker Configuration}
%\hfill \break
%\hfill \break
As described in the infrastructure chapter, the cluster being used can be configured with a different number of nodes and GPUs per node. %Distributed TensorFlow defines workers as the processes that perform training jobs, and each worker can claim a number of GPUs (or TPUs, or any other needed resource). 
In this experiment, we select the optimal layout with respect to the TensorFlow distributed training setup, including the number of GPUs per node, TensorFlow workers and number of GPUs in each worker. The objective is to understand the impact of communication overhead in the TensorFlow distributed setup.
The experiment has two sets of runs using a total of 32 GPUs.
In the first set the number of nodes (4) and the number of GPUs per node (8) are fixed. Different combinations of workers are selected: 32 workers with 1 GPU each, 16 workers with 2 GPUs each, 8 workers with 4 GPUs each and 4 workers with 8 GPUs each.
In the second set the number of nodes and GPUs per node varies: 32 nodes with 1 GPU each, 16 nodes with 2 GPUs each, 8 nodes with 4 GPUs each and 4 nodes with 8 GPUs each. The workers configuration follows the nodes configuration and is the same as in the previous run: 32 workers with 1 GPUs each, 16 workers with 2 GPUs each, 8 workers with 4 GPUs each and 4 workers with 8 GPUs each. Essentially, the same set of worker configurations is run on a different hardware layout. 
%Worth noting is that the last worker configuration in both runs (4 workers with 8 GPUs each) represents actually one run, since the hardware layout is the same in both runs.
%\begin{figure}%
%\setlength{\belowcaptionskip}{-15pt}
%    \centering
%    \subfloat\centering {{\includegraphics[width=6cm]{n_workers.png} }}
%    \caption{Impact of various combinations of hardware layout and TF worker configuration on time to perform one epoch of training. Orange represents the first set of runs, where hardware configuration is fixed with 4 nodes with 8 GPUs each, while number of workers and GPUs per worker change. Blue represents the second set of runs where number of workers and GPUs per worker correspond to number of nodes and GPUs per node.}
%    \label{kf-results-n-workers}
%\end{figure}
The results in figure  \ref{kf-results-batch-size}, on the right, show that the impact of different hardware configurations is not huge across the runs. The last three runs from both sets show times that differ by less than 3 seconds. The first run in both sets (32 workers with 1 GPU each) shows a significant increase in time, meaning the optimal configuration should not strive for less GPUs per node and less GPUs per worker. The best results are achieved when the number of workers matches the number of nodes and the number of GPUs per worker matches the number of GPUs per node. In subsequent experiments this setup will be used.

%\begin{figure}%
%\setlength{\belowcaptionskip}{-15pt}
%    \centering
%    \subfloat\centering {{\includegraphics[width=0.35\textwidth]{images/batch_times_32_workers.png} }}
%    \quad
%    \subfloat\centering {{\includegraphics[width=0.35\textwidth]{images/batch_times_4_workers.png} }}
%    \caption{Batch processing times for the second epoch of training using hardware layout with 4 %nodes with 8 GPUs per node. On the left side, batch processing times when number of workers is 32 and %each worker has 1 GPU. On the right side, batch processing times when number of workers is 4 and each %worker has 8 GPUs.}
%    \label{kf-results-n-workers-batch-times}
%\end{figure}
It was also observed that a higher number of workers introduces instability in the processing time per one batch, leading to instability in the processing time per one epoch and a less predictable time for the entire run completion. With a run with 32 workers, the variation in batch processing time is significant compared to the run with 4 workers. This goes in favor of avoiding higher number of workers and focusing on using nodes with as many GPUs as possible.

%\hfill \break
%\hfill \break
%\subsubsection{Total Number of GPUs}
%\label{sec:totalgpus}
%\hfill \break
%\hfill \break
We next determine how the performance changes with the increased number of GPUs available to workers. This set of runs strives for maximum utilization of the maximum number of GPUs available.
Building on the previous tests a batch size of 96 and one worker per node are selected, varying only the number of nodes and GPUs per node.
\begin{figure}%
\setlength{\belowcaptionskip}{-15pt}
    \centering
    \subfloat\centering %{{\includegraphics[width=0.45\textwidth]{total_gpu_linear.png} }}
    {{\includegraphics[width=0.45\textwidth]{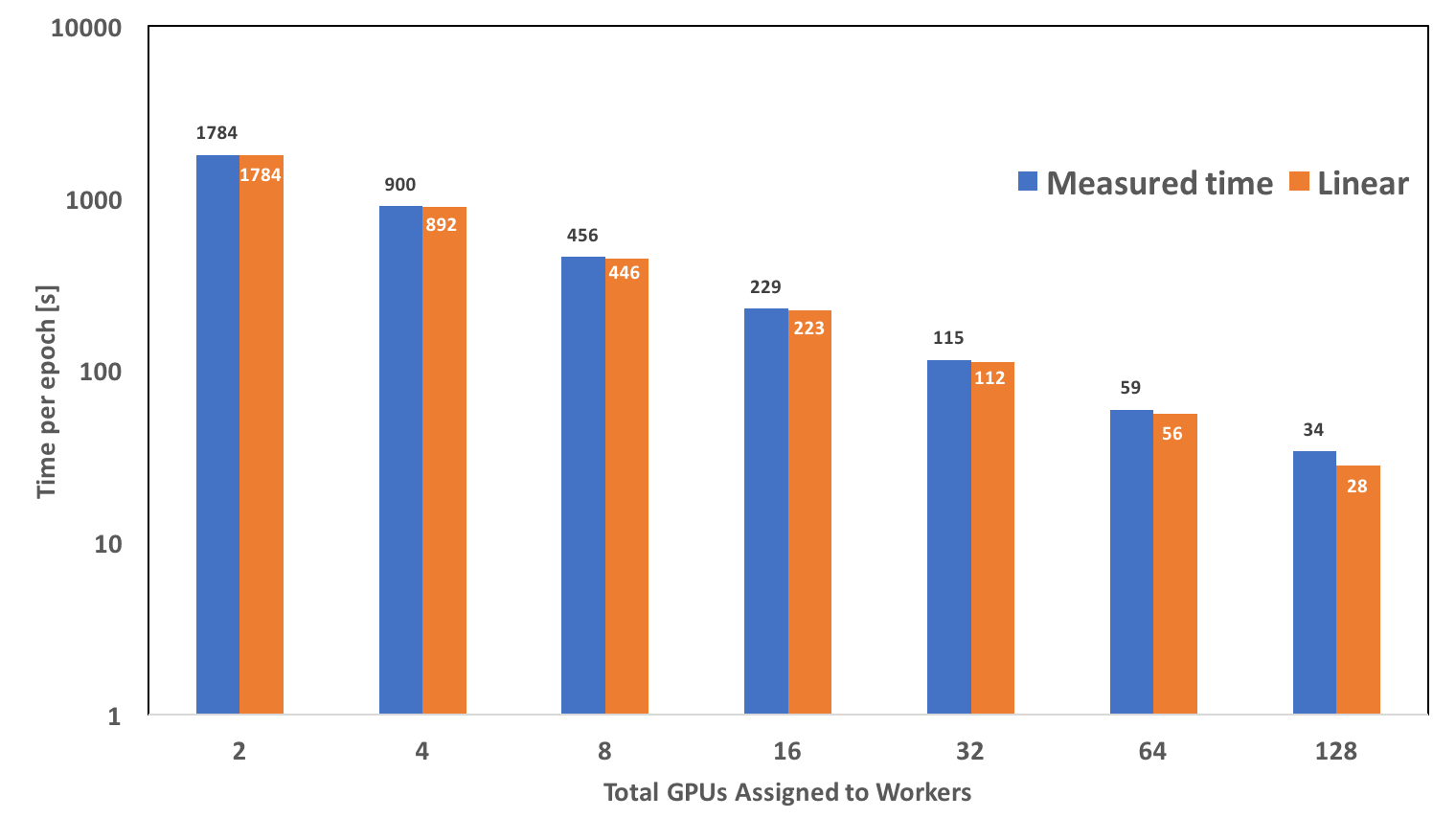} }}
    %\quad
    \subfloat\centering {{\includegraphics[width=0.45\textwidth]{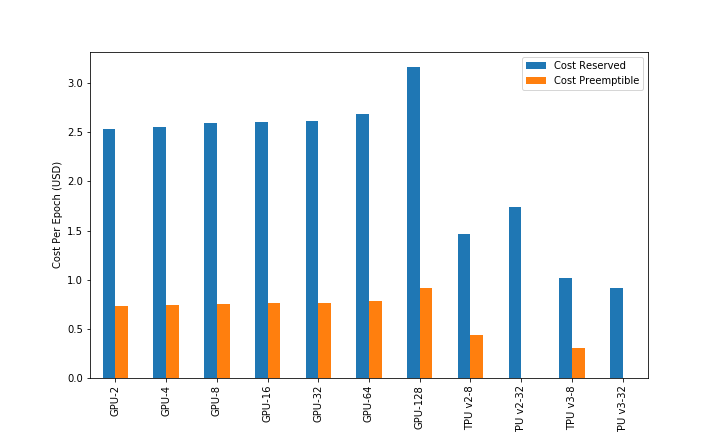} }}
    %\subfloat\centering {{\includegraphics[width=0.45\textwidth]{total_gpu_log.png} }}
    \caption{(left) The impact of increasing number of total GPUs assigned to workers on time to process one epoch of training. (right) Cost of running one epoch on the Google Cloud Platform. The cost per epoch remains similar for all GPU combinations, though the training time improves almost linearly.}
    \label{kf-results-n-workers-batch-times}
\end{figure}
The performance improvement is close to linear with a slight drop when moving to 128 GPUs, which are positive results regarding TensorFlow's distributed training capabilities to scale workloads horizontally (see figure \ref{kf-results-n-workers-batch-times}). This is particularly true when combined with Kubeflow and the ease of configuration and automation of the training process in large infrastructures.

\subsubsection{Cost Analysis Overview}
%\hfill \break

An important point when using public cloud resources is to understand upfront the costs involved in order to reach optimal service settings and configurations and, consequently, maximize cost effectiveness to reach a sustainable position.  %This section summarizes experiences evaluating the 3DGAN distributed training use case.
%\subsection{Google Cloud Platform}
%\hfill \break
The infrastructure costs during the tests on Google Cloud Platform are driven by the cost of the GPUs, with virtual machines accounting for less than 5\% of the total. Below, we present the numbers for using 2 to 128 GPUs and for a subset of TPU configurations. The calculation is done taking the training time of one epoch against the hourly GPU costs in region { \it europe-west4 } and include both reserved and preemptible options. It should be noted that when using reserved instances it is possible to apply for sustained or committed-use discounts which reduce the difference to preemptible instances costs.
%\begin{figure}%
%\setlength{\belowcaptionskip}{-15pt}
%    \centering
%    \subfloat\centering {{\includegraphics[width=6.5cm]{images/gcp_cost.png} }}
%    \caption{Cost of running one epoch on the Google Cloud Platform. The cost per epoch remains similar for all GPU combinations, though the training time improves almost linearly.}
%    \label{kf-results-batch-size}
%\end{figure}
A few points are worth noting (figure \ref{kf-results-n-workers-batch-times}, right panel): the cost per epoch remains similar when increasing the number of GPUs, while the training time is reduced up to 52 times for 128 GPUs (almost linear up to 64 GPUs); using preemptible GPUs can allow for significant cost savings and should be used whenever possible; finally, the best results are achieved using preemptible TPU v3-8, which are 2.4 times cheaper than their GPU equivalent, considering the training time. This is an interesting result that positions TPUs as a cost-effective option combined with their potential to reduce the overall training time. Preemptible TPUs  are only available up to 8 cores, with the reserved instance costs being less attractive. As an example, the overall training cost with 128 GPUs is the same as TPU v3-32, but it takes half the time. Future work will include a comparison with a larger number of TPU cores.

\subsection{Microsoft Azure with Azure's Machine Learning Service}
\label{sec:azure}
\begin{figure}%
    \centering
    \subfloat\centering {{\includegraphics[width=0.32\textwidth,  clip=true]{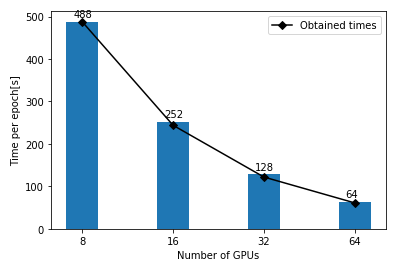} }}
    % \subfloat\centering {{\includegraphics[width=5.2cm]{images/azure_gpu100.pdf} }}
    \quad
     \subfloat\centering 
     {{\includegraphics[width=0.28\textwidth]{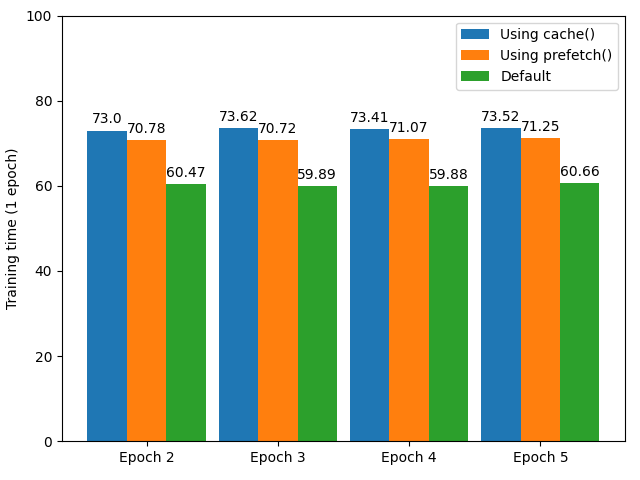} }}
    \caption{(left) Microsoft Azure training time per epoch for a $BS=64$ batch size. (right) Azure ML service automatic optimisation performance compared to manual cache and prefetch implementation. All times are measured in seconds.}
    \label{infiniband-comparison}
\end{figure}

\begin{figure}%
\setlength{\belowcaptionskip}{-15pt}
    \centering
    \subfloat\centering {{\includegraphics[width=0.35\textwidth]{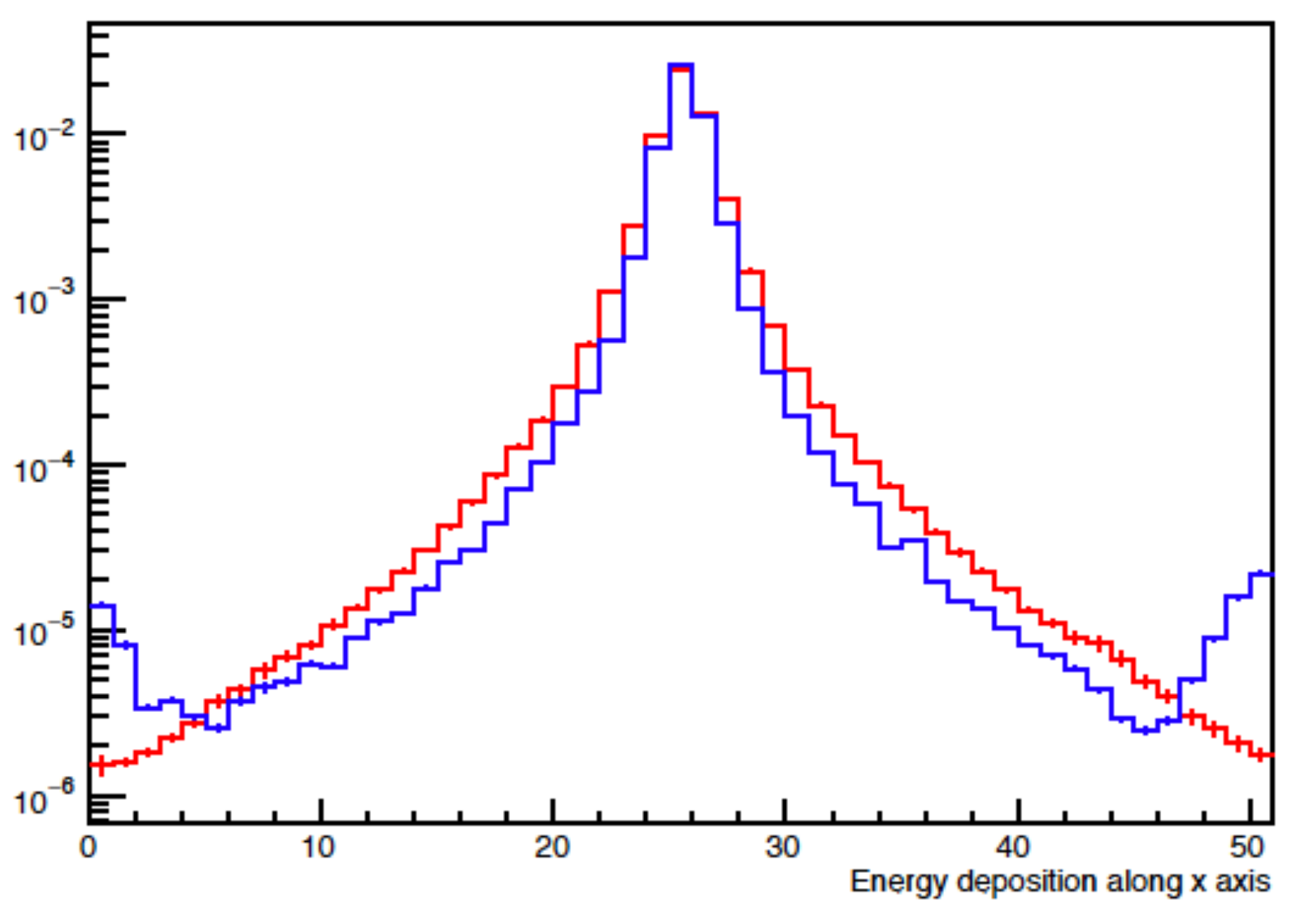} }}
    \qquad
    \subfloat\centering {{\includegraphics[width=0.35\textwidth]{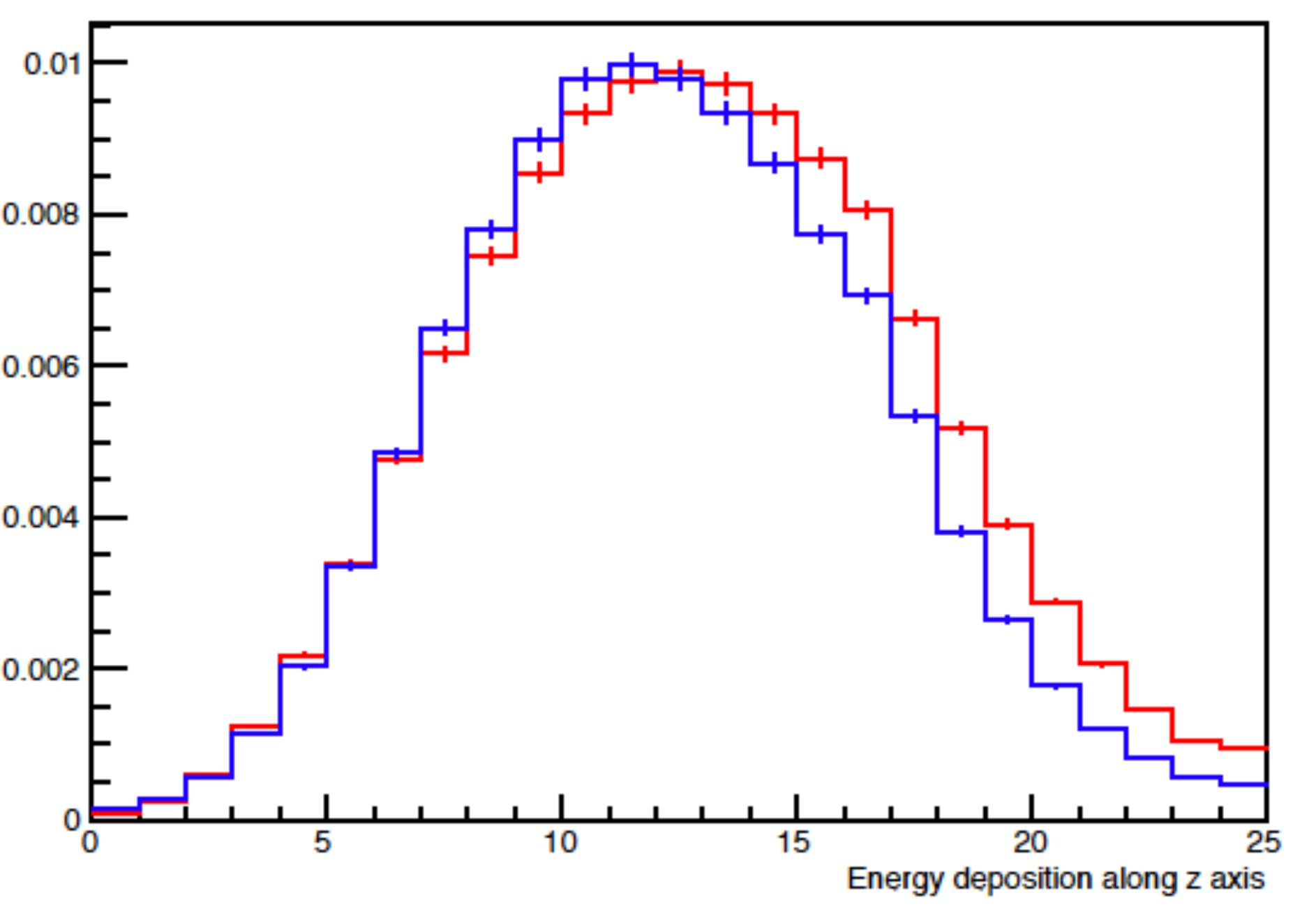} }}
    \caption{ Calorimeter energy response as predicted by 64 GPUs GAN  (blue) and Monte Carlo (red). (left) Distribution along the detector transverse plane in logarithmic scale.(right) Distribution along the calorimeter depth in linear scale. }
    \label{shapes_multinode}
\end{figure}

In the case of Microsoft Azure cloud, the 3DGAN deployment makes use of the Azure Machine Learning service \cite{azure} offered through CERN openlab's Azure cloud research enrollment. The service allows the user to be abstracted from the underlying infrastructure detail, since hardware is a managed component of the service stack. This means, essentially, that the job needs to be configured while the provisioning of the compute cluster is entirely operationalised by the Azure Machine Learning service, as part of the created deployment workflow.
%\subsubsection{Overview and architecture}
%\hfill \break
%\hfill \break
The interaction with the Azure ML service can be done using different ways such as Web interface, CLI and SDK. In this case, the SDK approach is preferred because of its wider support. It provides a space, called Workspace, where the user can manage the different run submissions. %When the user submits the run, the Azure ML Service will create a customised Docker image, to deploy the job on the compute nodes; it will compress the directory containing the training script and send it to the compute target, then mount the Workspace data store, containing the user data. 
A detailed description of the general architecture can be found in \cite{azure}.  %The general architecture is as depicted in Figure \ref{azureml-arch}, courtesy of Microsoft. 
%\hfill \break
%\begin{figure}%
%\setlength{\belowcaptionskip}{-15pt}
%    \centering
%    \subfloat\centering {{\includegraphics[width=\textwidth]{images/azureml_training-and-metrics.png} }}
%    \caption{Azure Machine Learning training and metrics.}
%   \label{azureml-arch}
%\end{figure}
%\subsubsection{Experiments and Results}
%\hfill \break
%\hfill \break
For our tests, we use GPU-powered nodes with 24 vCPU cores, 448 GiB memory and 4 V100 GPU.
%\begin{table}
%\begin{center}
% \begin{tabular}{||c c c c c||} 
% \hline
% Machine type & vCPU & Memory & V100  & RDMA \\ [0.5ex] 
% \hline\hline
% Standard\_NC24rs\_v3 & 24 & 448 GiB  & 4 & FDR rate \\ 
% \hline
% Standard\_NC24s\_v3 & 24 & 448 GiB & 4 & No \\ [1ex] 
% \hline
%\end{tabular}
%\end{center}
%\caption{Multi-GPU instances tested on Microsoft Azure cloud \cite{azure}}
%\label{table_azure}
%\end{table}
%The model is based on three-dimensional convolutions and therefore, it exhibits a high compute-to-communication ratio due to the cubic computational complexity of 3D convolutions operators: with a total number of parameters to be reduced that amounts to 5MB, the parameter transfer time is roughly 1 millisecond for a $50Gb/s$ network interconnect (FDR rate). As expected, the left panel on figure \ref{infiniband-comparison} shows no significant speed-up when training on a cluster with InfiniBand RDMA capable nodes, regardless of its size. Instead, 
The 3DGAN training time scales very close to linear up to a total of 16 nodes (64 GPUs). Azure ML automatically optimises the data set management, in aspects such as caching, pre-fetching and parallel data loading according to the hardware infrastructure setup. In figure \ref{infiniband-comparison}, right panel, we verify that manually modifying caching, pre-fetching and auto-tuning features in Tensorflow does not improve on the train time performance.
%{\it Standard\_NC24rs\_v3} instances are used in the following tests.
In terms of image fidelity, the work in \cite{IXPUG_3DGAN} has shown that data parallel training can affect the quality of the images especially at the border of the detector sensitive region. Indeed, in figure \ref{shapes_multinode}, on the left, the calorimeter energy response in the detector transverse plane shows a similar performance degradation at the edges of the energy distribution. The amount of energy deposited in this region is orders of magnitudes smaller than in the central pixels; it is therefore much harder for the GAN to correctly predict it. It should be noted, however, that this behaviour does not critically affect physics applications, since such small energy deposits are usually neglected. On the other hand, the slight shift of the longitudinal energy distribution, on the right in figure  \ref{shapes_multinode}, is significant and additional optimisation is needed in order to reduce this effect.

The experiments on The CERN openlab Azure enrollment were performed using an Azure credits grant, %This grant has limited options to access types of instances (e.g. spot vs. reserved) or even the most effective cloud regions to access GPU quotas at scale (hundreds). 
currently evolving to provide capabilities for realistic analysis and overall resource costing optimisation. These capabilities were not yet ready at the time this paper was produced and for this reason we defer any cost analysis to future studies.
\section{Conclusions and Future Plans}
With Deep Learning models in the HEP domain becoming more complex,  computational requirements increase, triggering the need to consider HPCaaS and MLaaS offers in the public cloud. 
This work presents results of the first deployment of a three-dimensional convolutional GAN for detector simulation on TPUs, demonstrating an efficient parallelization of the adversarial training process: the 3DGAN training time is brought down from about a week to around one hour. This result enables large architecture hyper-parameter scans to run in just a few days (instead of weeks) and, therefore, it greatly extends the range of detector geometries than can be simulated by the 3DGAN model. %From a physics perspective, we verify that low precision data representation (bfloat16) does not deteriorate the synthetic image fidelity and we find minimal performance degradation once we increase the level of parallelism of the training process. 
Validation against Monte Carlo shows a slight over-estimation of the energy distribution along the edges of the detector sensitive volume for a large number of nodes (above 64 GPUs): simple strategies, typically employed to ease the effect of increasing global batch sizes are not sufficient to improve convergence at scale. We are currently investigating several different directions, such as the role played by the batch normalisation layer, which is one of the components greatly affecting the 3DGAN model performance \cite{GAN_Gulrukh}.
%In addition we plan to extend the scope of this work in two main directions: data sample size (i.e.larger number of detector read-out channels) and applications. 3DGAN is, indeed, representative of a simulation workload, while we plan to investigate different architectures (such as graph neural networks) in order to provide a comprehensive overview of HEP DL use cases.
In addition to the physics result, this work demonstrates the deployment of scientific DL workloads using public cloud services, complementing the  on-premises infrastructure offers in use by research organisations. This is achieved by exploring innovative, open hybrid provisioning and orchestration models, profiting from the technological potential of each cloud provider, unlocked from any particular commercial vendor or provisioning model.
Although public cloud services are offered in the general market as commodity services, they must be validated for research use cases  technically but also in terms of cost optimisation requirements.  These elements combined, as part of a global strategy when scaling out on-demand heterogeneously, can profit from the full potential of cloud-based services that are evolving from basic elastic provisioning of virtual resources, to a transparent and adaptive smart continuum.
%, making a plethora of innovative workflows available for research data intensive applications.
These possibilities must be assessed with close alignment across performance, usability and economics, considering complexities such as type of resources, data movement patterns, optimisation of architectures adapted to the scientific workloads, provisioning lock-in risks in and also data processing aspects to achieve a long-term sustainable position. %, at the same time as the baseline of data science requirements. These aspects must be considered integrated and interdependent, as part of an end-to-end workflow to achieve a long-term sustainable position.
Future work will continue to explore models of resources combination from specific vendors  and explore seamless orchestration of resources federated across multiple cloud providers, combining technical and cost-effectiveness requirements. In addition, data governance aspects will also be considered, ensure the presence of a set of technical measures to guarantee adequacy with current legislation (e.g. GDPR and Free Flow of data).

%
% BibTeX or Biber users please use (the style is already called in the class, ensure that the "woc.bst" style is in your local directory)
% \bibliography{name or your bibliography database}
\bibliography{mybibliography.bib}
%\begin{thebibliography}{}
%Journal Author, Journal \textbf{Volume}, page numbers (year)
%Book Author, \textit{Book title} (Publisher, place, year) page numbers
% etc
%\end{thebibliography}

\end{document}